\documentclass[preprint,12pt]{elsarticle}

\usepackage{amssymb}
\usepackage{amsmath}

\usepackage[toc,page]{appendix}
\renewcommand\appendixname{Appendix}

\usepackage[linesnumbered,ruled,vlined]{algorithm2e}
\SetKwInput{KwInput}{Input}
\SetKwInput{KwOutput}{Output}

\usepackage{xurl}

\usepackage{booktabs}

\biboptions{square,numbers}

\begin{document}

\begin{frontmatter}


\title{Classifying the evolution of COVID-19 severity on patients with combined dynamic Bayesian networks and neural networks}

\author[upm]{David Quesada}
\author[upm]{Pedro Larrañaga}
\author[upm]{Concha Bielza}

\address[upm]{Departamento de Inteligencia Artificial, Universidad Politécnica de Madrid, Madrid, Boadilla del Monte, 28660, Spain}

\begin{abstract}
When we face patients arriving to a hospital suffering from the effects of some illness, one of the main problems we can encounter is evaluating whether or not said patients are going to require intensive care in the near future. This intensive care requires allotting valuable and scarce resources, and knowing beforehand the severity of a patients illness can improve both its treatment and the organization of resources. We illustrate this issue in a dataset consistent of Spanish COVID-19 patients from the sixth epidemic wave where we label patients as critical when they either had to enter the intensive care unit or passed away. We then combine the use of dynamic Bayesian networks, to forecast the vital signs and the blood analysis results of patients over the next 40 hours, and neural networks, to evaluate the severity of a patients disease in that interval of time. Our empirical results show that the transposition of the current state of a patient to future values with the DBN for its subsequent use in classification obtains better the accuracy and g-mean score than a direct application with a classifier.
\end{abstract}


\end{frontmatter}

\pagestyle{plain}
\section{Introduction}

Throughout the COVID-19 pandemic, healthcare systems all around the world have suffered a staggering pressure due to the sheer number of infected patients that arrived to medical centers. The nature of this pandemic was such that patients could range from completely asymptomatic to presenting critical respiratory issues. As such, and given that the amount of resources in medical centers is limited, it was a crucial task to discern whether or not a patient presented symptomatology that could devolve into a critical condition or into only mild afflictions.

The issue of predicting the clinical outcome of COVID-19 patients has seen much interest in recent years. Some authors opted for discerning the severity of the illness depending on certain comorbidities like heart failure \citep{arevalo2022comorb}, neurodegenerative diseases \citep{yu2021alzheimer}, cardiovascular diseases \citep{ehwerhemuepha2021super}, or chronic pulmonary diseases \citep{momeni2021dynamic}. These studies have shown that comorbidities related to COVID-19 increase the risk of death of a patient. As such, many efforts are also put into preprocessing clinical data and selecting an appropriate set of variables that define the effect of the illness.

From the point of view of predicting the outcome from data, many machine learning approaches have been tested in the literature. Some authors opted for performing a statistical analysis and applying logistic regression for classifying mortality \citep{yu2021alzheimer,wu2020development,xiong2020pseudo,berenguer920logistic}. Another popular approach consists of training simple perceptron or multilayer neural network models to approximate a function that relates the variables in the system and classifies patient instances \citep{pinter2020covid,dhamodharavadhani2020covid,aznar2021clinical,kianfar2022spatio}. Tree-based models like random forests \citep{aznar2021clinical,cornelius2021rf,pourhomayoun2021rf,tezza2021rf} or XGBoost \citep{aznar2021clinical,bertsimas2020xgb,vaid2020xgb,yadaw2020xgb} are also some of the most popular and best performing tools for this task. In the case of interpretable models, Bayesian networks have also been applied to predicting the severity of COVID-19 on patients while also trying to gain some insight on the problem at hand \citep{fenton2021bayesian,vepa2021bn}

Another possible approach is to view the problem as a time series forecasting issue. Each patient that arrives at a hospital has its vital signs measured and has blood analysis performed on them. Afterwards, if the patient is not discharged and requires further care, new recordings are performed on a semi-regular basis. This generates time series data of each patient, where measurements are taken over a period of several hours each until either the patient overcomes the illness or passes away. In this scenario, time series models can be applied to forecast the state of a patient and predict whether they will be suffering from severe symptoms in the near future or not. This approach has also been explored in the literature with models like dynamic Bayesian networks \citep{pezoulas2021ts}, recurrent neural networks \citep{villegas2021ts} and dynamic Markov processes \citep{momeni2021ts}.

In this work, we took a hybrid approach between static and dynamic models. We used data recovered from patients infected with the sixth Spanish COVID-19 wave that arrived to the Fundación Jiménez Díaz hospitals in Madrid. After preprocessing this data and selecting an appropriate variable set, we trained hybrid models between dynamic Bayesian networks (DBN) as forecasting models and neural networks (NN) as classifier models. The main idea of our proposal is to obtain the first vital signs and blood analysis from a patient and then perform forecasting of these variables with the DBN model up to a certain point in the near future. Afterwards, we can use the classifier model to identify the forecasted values as critical or not critical. This procedure can help identifying whether a patient that just arrived to triage in a medical center is going to worsen significantly in the following days. 

The rest of this paper is organized as follows. Section \ref{sec:bg} gives some background on dynamic Bayesian network models. Section \ref{sec:classDBN} explains the architecture of the hybrid model with the neural network, where this classifying model is interchangeable with any other static classifier. Section \ref{sec:results} shows the experimental results of the tested models. Finally, Section \ref{sec:concl} gives some conclusions and introduces future work.

\section{Dynamic Bayesian networks}\label{sec:bg}

Dynamic Bayesian networks \citep{koller2009} are a type of probabilistic graphical model that represent conditional dependence relationships between variables using a directed acyclic graph. They extend the framework of Bayesian networks to the case of time series. Similarly to static BNs, each of the nodes in the graph represents a variable in the original system and the arcs represent their probabilistic relationships. In the case of DBNs, time is discretized into time slices that represent consecutive instants. This way, we have a representation of all the variables in our system across time. Let $\mathbf{X}^t = \{ X^t_0, X^t_1, \ldots, X^t_n \} $ be the set of all the variables in the time slice $t$. Then, we can define the joint probability distribution of the network up to some horizon $T$ as:

\begin{equation}
p(\mathbf{X}^0, \ldots, \mathbf{X}^T) \equiv p(\mathbf{X}^{0:T}) = p(\mathbf{X}^0) \prod_{t=0}^{T-1} p(\mathbf{X}^{t+1} | \mathbf{X}^{0:t}),
\label{eq:dyn_jpd}
\end{equation}

where $p(\mathbf{X}) = \prod_{i=0}^{n} p(X_i|\mathbf{Pa}_i)$ represents the probability distribution of a set of nodes $\mathbf{X}$ and $\mathbf{Pa}_i$ represents the set of parent nodes of $X_i$ in the graph. However, in Equation \ref{eq:dyn_jpd} all time slices $\mathbf{X}^{0:T}$ have to be taken into account to calculate the joint probability distribution. In this scenario, it is very common to assume that the future state of the system is independent of the past given the present. A DBN that follows this assumption is called a first-order Markovian network. This implies that only the last instant is used to calculate the next one and it simplifies the calculation of the joint probability distribution greatly:

\begin{equation}
p(\mathbf{X}^{0:T}) = p(\mathbf{X}^0) \prod_{t=0}^{T-1} p(\mathbf{X}^{t+1} | \mathbf{X}^{t}).
\label{eq:markov_1}
\end{equation}

\begin{figure}[!t]\centering
	\includegraphics[width=0.6\columnwidth]{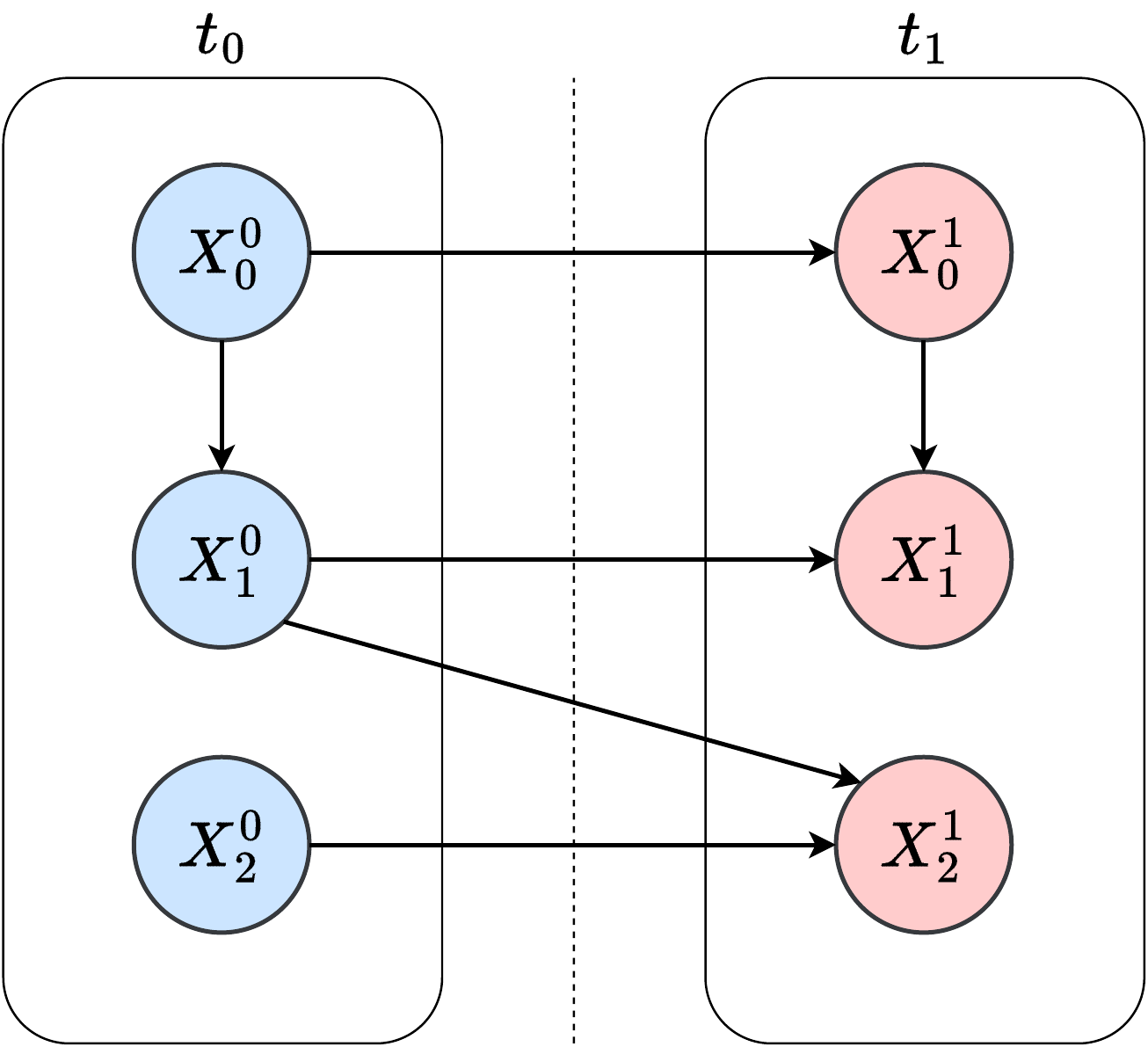}
	\caption{Example of the structure of a first-order Markovian DBN with two time slices $t_0$ and $t_1$. To calculate the future values in $t_1$, we would only need to know the current values of our variables in $t_0$.}
	\label{fig:dbn_fig}
\end{figure}

An example of the structure of a DBN with Markovian order 1 is shown in Fig. (\ref{fig:dbn_fig}). One advantage that DBN models present is that they do not need to be trained with time series of constant length. Due to the Markovian order assumption in Equation (\ref{eq:markov_1}), we only need to recover several batches of two consecutive instants from the original dataset to learn the structure and parameters of the network. We can use several time series with different lengths recovered from the same stochastic process to train a DBN model from data. The reason for this is that we only need the values of the variables inside the temporal window defined by the Markovian order to train our model, so the total length of the time series is not relevant in the learning phase. This helps when applying this kind of model to real-world problems, where the length of the data from processes can vary depending on circumstances outside of the system.

\section{Combining DBNs and static classifiers}\label{sec:classDBN}

When we predict COVID-19 severity on patients in the near future, we face several issues. On one hand, we only have the data of their vital signs and blood analysis when the patient first arrives at the hospital. As we are interested on their state on the following days, we need to forecast the evolution of these variables over time. On the other hand, we need a mechanism that identifies given a state vector of a patient whether they are in a critical state or not.

\subsection{Forecasting the state vector}

When a patient afflicted with COVID-19 stays in intensive care for a prolonged period of time, they are monitored and new readings of their vital signs and blood analysis are recorded on a semi-regular basis of several hours. All the variables in these instances form a state vector $\mathbf{S} = \left[s_0, s_1, \ldots, s_n\right]$ at each point in time, and the final data recovered from a patient $k$ is a vector of instances $\mathbf{P}_k = \left[ \mathbf{S}^0, \mathbf{S}^1, \ldots, \mathbf{S}^T \right]$ ordered in time from the oldest vital sign readings and blood analysis to the most recent ones. When we combine several patients data, it generates a time series dataset that can be used to train a time series forecasting model. It is worth noting that the length $T$ of the data from each patient depends on the time they spent in the hospital. If a patient is discharged with only one vital sign reading and blood analysis, then we do not have data with a time component. In this situation, this patient could not be used for training our temporal model.

Given that in our case all the variables in a state vector $\mathbf{S}^t$ are continuous, we will use a Gaussian DBN to model the dependencies and to perform forecasting. A DBN model can help us gain some insight on which variables have a greater impact on the evolution of a patient. Furthermore, the ability of DBNs to be trained with different length time series after deciding a Markovian order is also relevant in this problem, given that the number of instances per patient varies greatly. By setting a Markovian order 1, we will be able to use the data from all patients except the aforementioned ones with a single reading, where no temporal data at all can be used.

After training the DBN model, we can use it to forecast the state vector of a patient up to a certain point in the future. This forecasting represents an estimate of the evolution that the patient will undergo, and it can be used to assess whether it will lead to severe symptoms or not. This process effectively gives an estimate of the future vital signs and blood analysis of a patient without spending additional resources and time on it. 

\subsection{Classifying critical values}

The task of evaluating whether a patient is in a critical state of the COVID-19 infection has been performed in the literature mainly through some kind of medical score \citep{fan2020score} or by labelling instances due to some external indicator, for example being transferred to the intensive care unit. If we obtain a labelled dataset of patients through any of these methods, we can then take a machine learning approach by training classifier models that identify whether a patient is in a critical state given their state vector $\mathbf{S}$.

\begin{figure}[!t]\centering
	\includegraphics[width=\columnwidth]{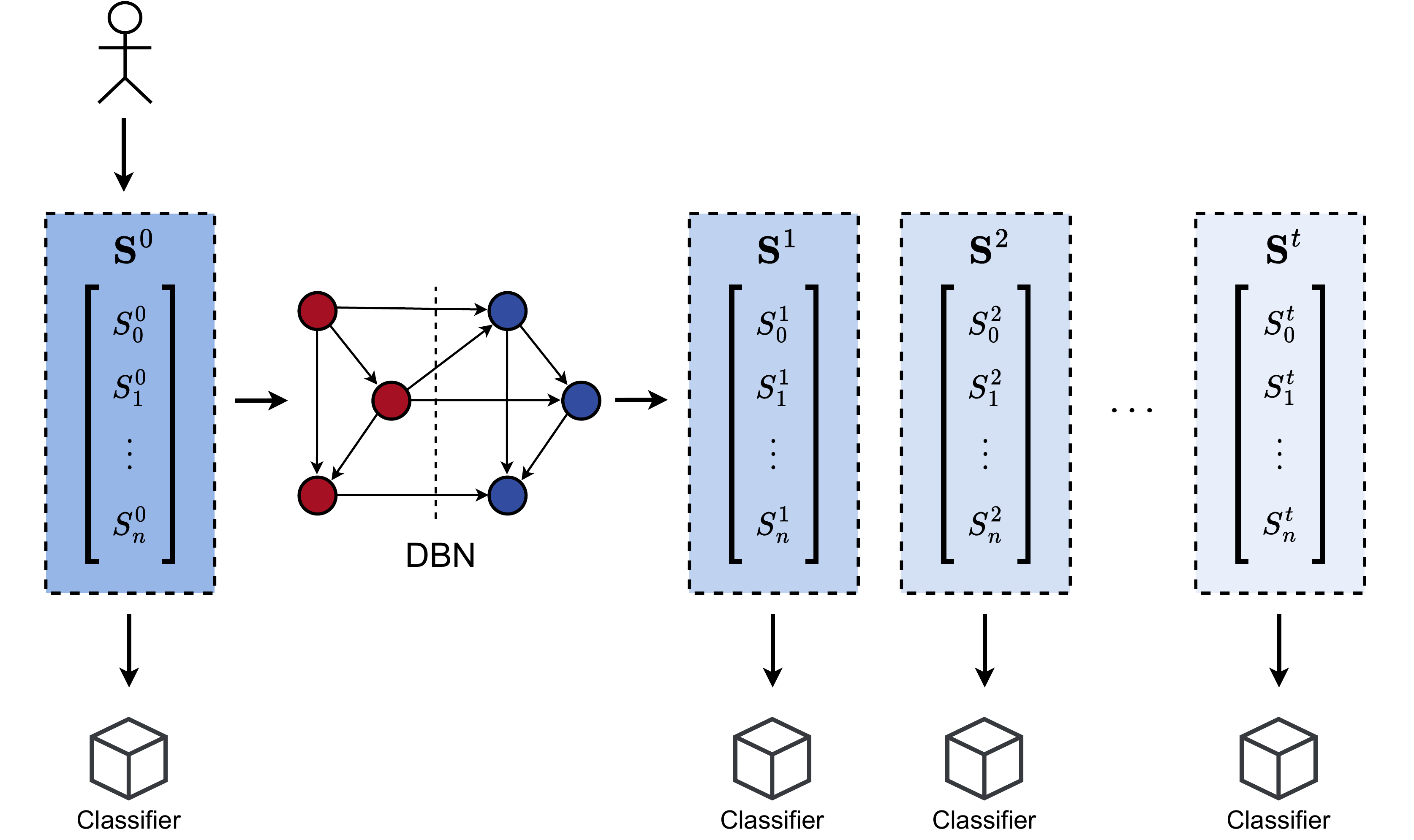}
	\caption{Schematic representation of the classifier-DBN framework. After obtaining a state vector $\mathbf{S}_0$ from a patient, we can use it to forecast the next $t$ state vectors with the DBN model and check if they are critical with our static classifier.}
	\label{fig:hybrid_fig}
\end{figure}

If we combine this approach with the forecasting of the state vector, we get a hybrid model between static classifiers and time series models that is capable of evaluating the present and near future condition of a person suffering from COVID-19. When a patient arrives at a hospital and gets their vital signs and blood analysis recorded, we obtain the state vector $\mathbf{S}^0$ of the very first instant of time. Then we can feed $\mathbf{S}^0$ to a trained classifier model to evaluate whether this patient is already in a critical state or not. If this is not the case, we can then use $\mathbf{S}^0$ as the starting point for our DBN to perform forecasting. This will return us the values of $\mathbf{S}^1, \mathbf{S}^2, \ldots, \mathbf{S}^t$ up to a certain point $t$ in time. All these state vectors can in turn be classified to evaluate the expected severity of the symptoms in that patient. With this method, we can see if a patient is expected to end up suffering from critical COVID-19 and when approximately will this situation occur. To illustrate this whole process, a schematic representation of this framework can be seen in Fig. (\ref{fig:hybrid_fig}).

Our proposed framework supports any kind of classifier that is able to produce a discrete prediction given a continuous state vector $\mathbf{S}^t$. We used a modular implementation where the classifier used can be a support vector machine, an XGBoost, a neural network and a Bayesian classifier. All these classifiers have seen use in the literature and could find applications where one is more effective than the others. Due to this architecture, any other classifier model could potentially be introduced as a new module if the need arises.

\begin{figure}[!t]\centering
	\includegraphics[width=\columnwidth]{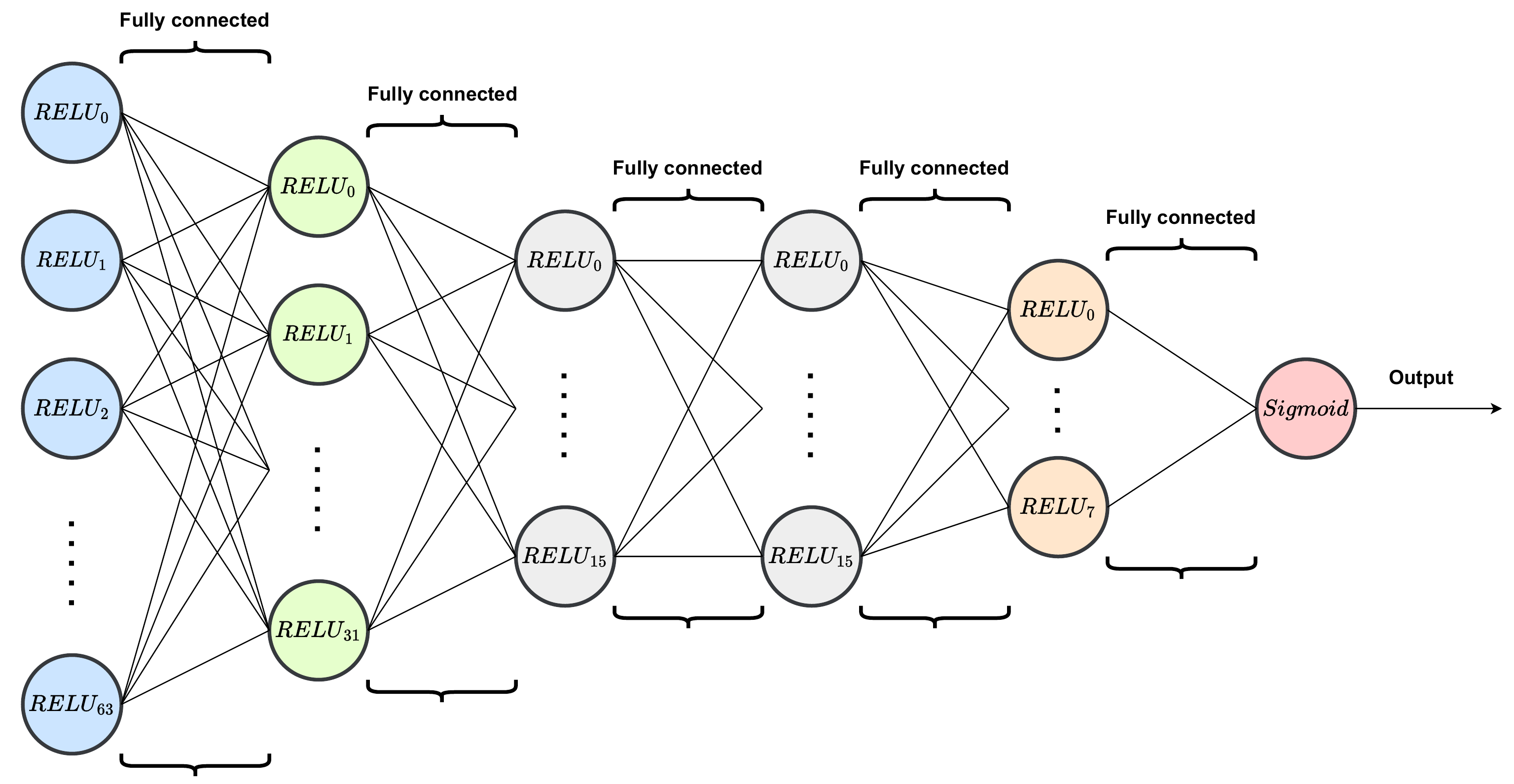}
	\caption{Structure of the neural network model used in the experiments. }
	\label{fig:nn_arch}
\end{figure}

In our case, the architecture that was most effective was the combination with a neural network. The network had an internal structure of 5 hidden dense layers with 64, 32, 16, 16 and 8 neurons each. They all used RELU activation functions and had their weights initialized with the identity. The last layer used a single neuron with a sigmoid activation function for binary classification. A result greater than 0.5 is equated to predicting a critical status for a patient, and a result lesser or equal to 0.5 predicts a non-critical scenario. A representation of this structure can be seen in Fig. (\ref{fig:nn_arch}).

\section{Experimental results}\label{sec:results}

For our experiments, we used a dataset consisting of anonymous data recovered from 4 different Spanish hospitals from the Fundación Jiménez Díaz in Madrid. After preprocessing it, we used this data to fit our proposed model and evaluate its capabilities to predict the future critical status of patients suffering from COVID-19 infections.

\subsection{Preprocessing}

Our raw dataset covers the period from the 27th of October 2021 to the 23rd of March 2022. In total, there are 21.032 rows with incomplete data from 15.858 patients and 532 variables, most of which present missing values for the majority of patients. This is a common occurrence in a medical dataset of these characteristics, given that not the same tests are performed to all the patients and some of the results have to be recorded manually. This data covered patients that had confirmed cases of COVID-19 via a positive PCR test.

The consecutive rows in the dataset that correspond to a same patient are ordered in time forming time series sequences. However, the frequency at which the instances were recorded is uneven. This is due to the fact that performing blood analysis from patients and obtaining the results does not take a fixed amount of time and is not always performed after fixed intervals. To tackle this issue, we established a period of 4 hours between each row and formed batches of instances where missing data was filled with the average values of the rest of instances in the same batch. This 4 hour period was chosen because usually new tests were performed on average roughly after every 4 hours in our dataset.

\begin{figure}[!t]\centering
	\includegraphics[width=\columnwidth]{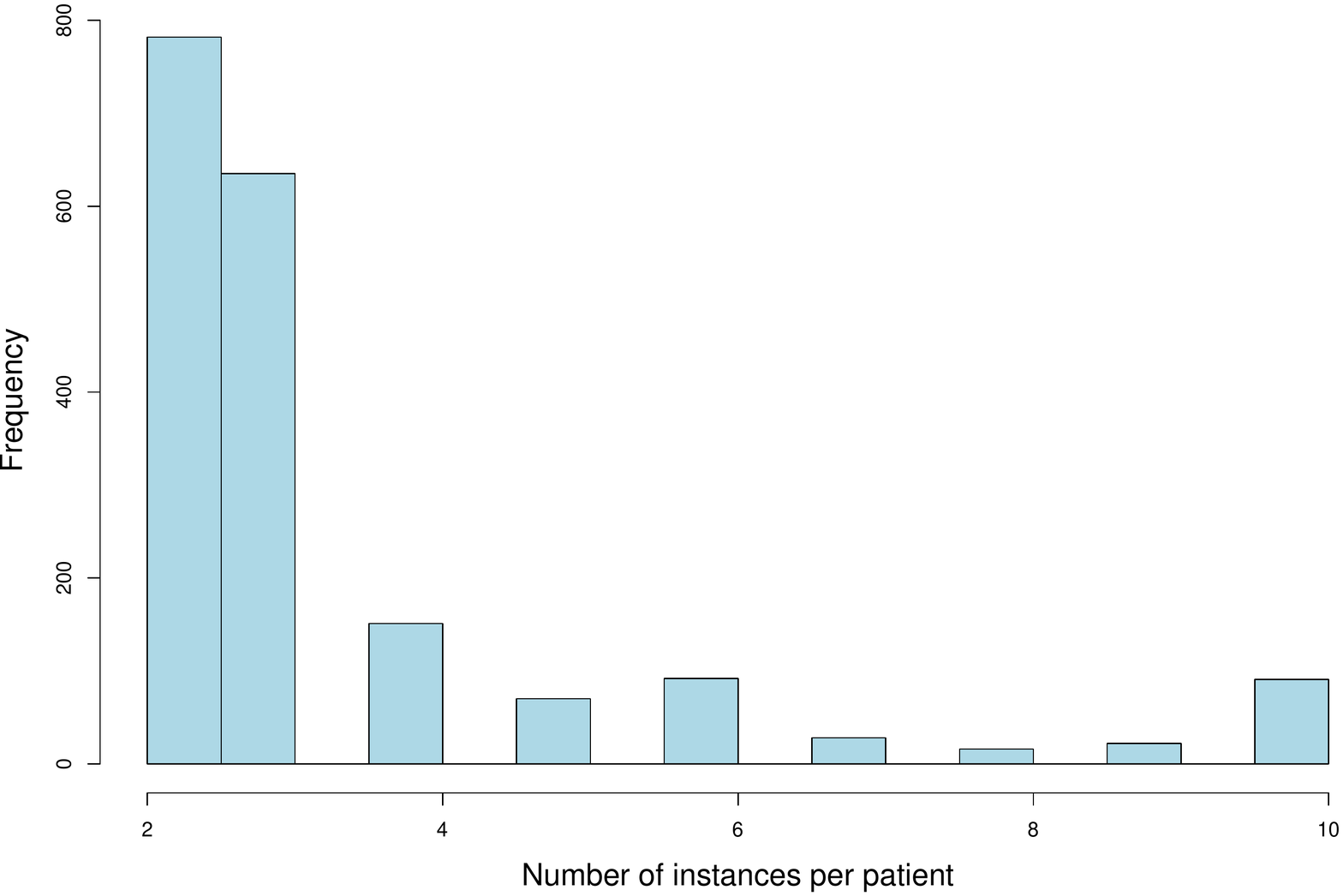}
	\caption{Histogram with the number of instances per patient greater than 1 in the dataset. Inside the last bracket we have grouped all the patients with 10 or more instances. A higher number of instances indicates a longer stay in the hospital and as such a more severe case of COVID-19, which is far less common than a mild case.}
	\label{fig:hist}
\end{figure}

From the 21.032 rows, 13.971 were from patients that appear only in a single instance, where the vast majority were discharged from the hospital afterwards due to mild symptomatology and only 48 of these patients passed away. This data cannot be used to train the DBN models, given that a single register is not enough to form a time series sequence. However, it will be used to train the classifier models. From the remaining patients with more than a single instance, the majority of them have either two or three rows of recorded values. To illustrate this, we show a histogram with the distribution of the number of instances per patient in Fig. (\ref{fig:hist}).

Regarding the 532 variables in our dataset, most of them correspond to specific values in uncommon tests and analysis, and they have over 70\% of missing values across all instances. In our case, we have opted for reducing the number of variables to only those that are obtained from the vital signs of a patient, like their body temperature and their heart rate, their descriptive characteristics like age, gender and body mass index, and the variables from a regular blood analysis like the albumin and D-dimer values. All these variables are routinely taken when a patient arrives at urgent care and obtaining them does not pose a severe expense of resources. This reduced the number of variables to 62, and from those we chose to retain the vital sign readings and the descriptive characteristics, while allowing feature subset selection on the blood analysis related variables. This subset selection was performed via random forest importance on classification on our objective variable, which will be whether or not a patient was put in the intensive care unit or passed away. This is what defines our critical cases of COVID-19, which are only a 18.8\% of the total number of patients in our dataset.

\subsection{Experiment results}

In this section we show the experimental results obtained with different combinations of classifier-DBN models. For our experiments, we used an XGBoost, a support vector machine, a neural network and a Bayesian classifier. In particular, this Bayesian classifier is a tree-augmented naive Bayes built following the hill climbing super-parent (HCSP) algorithm \citep{keogh2002learning}. All the project was coded in R and is publicly available online in a GitHub repository\footnote{https://github.com/dkesada/Class-DBN}. The dataset used is not made public due privacy and legal reasons.

Regarding the software we used in our experiments, the DBN models where trained using our own public package ``dbnR''\footnote{https://github.com/dkesada/dbnR}, the XGBoost models where trained with the ``xgboost'' package \citep{chen2022xgboost}, the support vector machines where trained with the ``e1071'' package \citep{meyer2022e1071}, the neural networks with the ``keras'' R interface \citep{allaire2022keras} and the Bayesian classifiers were trained with the ``bnclassify'' package \citep{mihaljevic2018bnclassify}. The parameters of each classifier were optimized using differential evolution with the R package ``DEoptim'' \citep{mullen2011deoptim} based on the geometric mean (g-mean) \citep{tharwat2021classification} of the models. This metric is defined as $g_m = \sqrt{recall * specificity}$, which uses all values in the resulting confusion matrix when calculating the final score. Using both the recall and the specificity of the predictions ensures that the imbalance between critical cases and non-critical cases is taken into account when optimizing the parameters. We do not want a model optimized solely on accuracy because it would lead to models that only predict the majoritary class of non-critical for all patients.

To alleviate the issue of imbalanced data, we also applied SMOTE oversampling with the `DMwR' package \citep{chawla2002smote, torgo2010dmwr} to synthetically generate instances of both critical and non-critical cases. This is a common practice that creates synthetic data to offset the difference between the number of instances of the majority and minority classes. In our case, we will use SMOTE to create modified datasets for training our classifiers. This will help the models to avoid getting stuck on predicting the majority non-critical class for almost all instances.

To test our hybrid models, we take the state vector of a patient in an instance and forecast up to 10 instants into the future with the DBN model. Then, we use the classifier model to classify each of this forecasts as critical or not and we compare the predicted label with the true label of the instance. Given that each instance is separated from the next one by 4 hours, in total we forecast 40 hours into the future with the DBN model. With this method, we will be able to see the behaviour of the classifiers and the changes in accuracy and g-mean as we use state vectors from further into the future. The average results obtained across all forecasts of the models can be seen in Table \ref{tb:avg_res}.

\begin{table}[!htb]
\resizebox{\columnwidth}{!}
{
\begin{tabular}{lcccc}
 \toprule
 & \bfseries Accuracy & \bfseries g-mean & \bfseries Train (h) & \bfseries Exec (s)  \\ 
 \midrule
 XGBoost & 0.698 & 0.455 & 1.950 & 9.634  \\  
 SVM & 0.735 & 0.522 & 1.145 & 9.654  \\  
 NN & 0.771 & 0.541 & 1.384 & 9.863  \\  
 HCSP & 0.736 & 0.468 & 1.046 & 9.878  \\
 \bottomrule
\end{tabular}
}
\caption{Mean results in terms of the accuracy, g-mean score, training and execution time of the models on average for all the experiments. It is worth noting that training time includes optimization of parameters, which involves the creation of multiple models to evaluate different configurations.}
\label{tb:avg_res}
\end{table}

The results in Table \ref{tb:avg_res} show that, on average, the most accurate model is the neural network in both accuracy and g-mean. The performance of both the SVM and the HCSP are very similar in terms of accuracy, but the difference in g-mean score of the SVM shows that it is able to discern better the more uncommon critical instances. For this particular case, although the XGBoost model is very popular in the literature, it obtains worse overall results than the rest of the classifiers. In our experiments, due to the imbalance between classes we had to find a compromise between the global accuracy and the accuracy of the minoritary class. If left unchecked, the models would become biased to the majoritary class and predict almost unanimously every single instance as non-critical, invalidating the use of the model while obtaining accuracies close to 90\%. By using the g-mean as optimization metric in combination with the SMOTE oversampling, we were able to alleviate this problem. A high accuracy on the majoritary class of non-critical patients will be able to help reduce the oversaturation of ICU resources, given that all models can evaluate whether a patient will reach a critical state of the COVID-19 infection or not in less than 10 seconds. On the other hand, being able to discern the few critical cases that arise is also needed to help doctors determine which patients need more specific care to try to reduce the mortality rate. On the topic of training time, training and tuning the models takes on average between one and two hours. Given that these kind of models should not need to be retrained until some significant issue happens with the disease, like a new variant or new specific symptoms appear on patients that differ from the training data used, this training times should be reasonable to be performed once.

\begin{figure}[!t]\centering
	\includegraphics[width=\columnwidth]{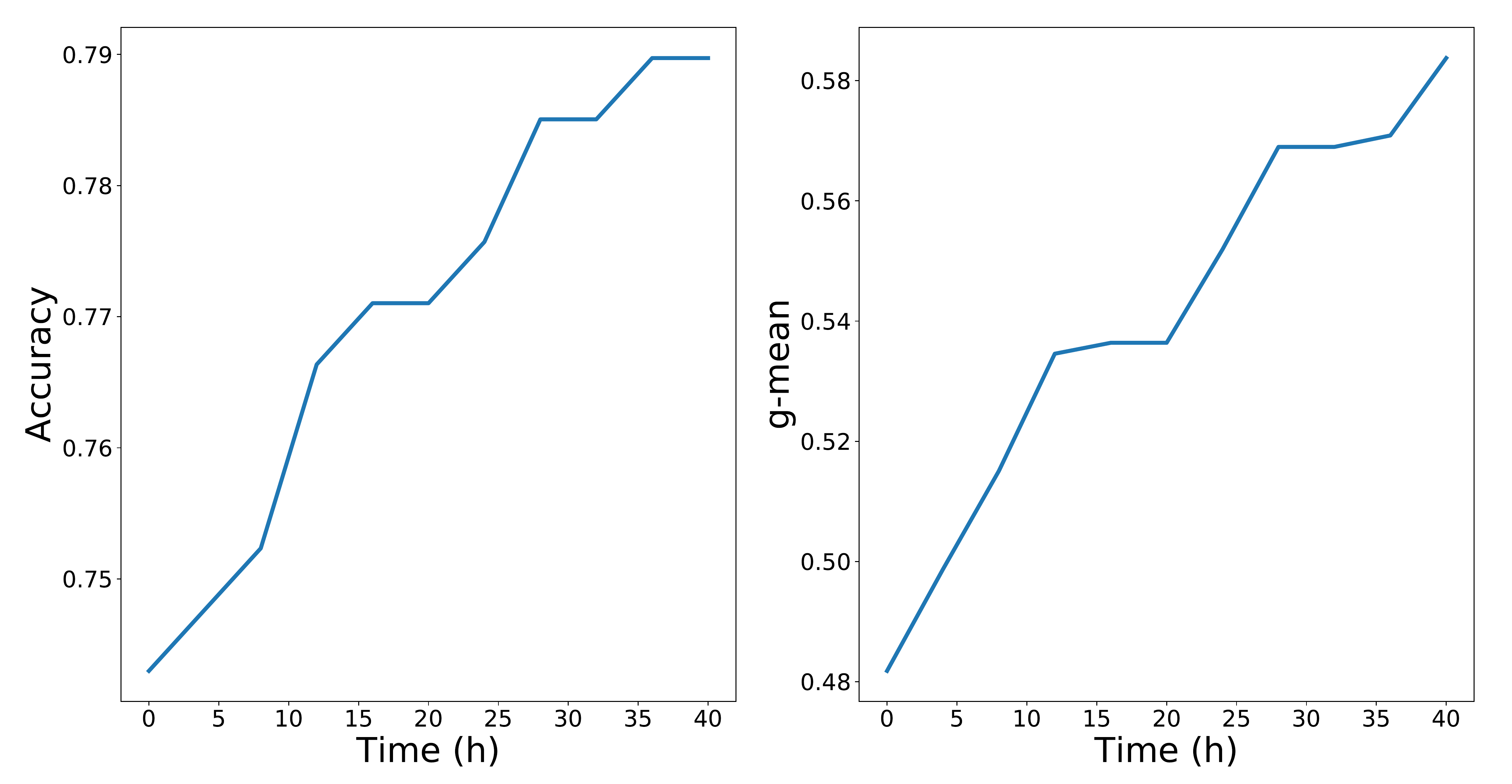}
	\caption{Classification results of the neural network model as we feed it state vectors further ahead in time with the DBN model. The classification performance of the neural network improves monotonically by combining it with the DBN forecastings.}
	\label{fig:subplots}
\end{figure}

Given that the model with the NN obtains the best average results, we show in Fig. (\ref{fig:subplots}) the details of its performance depending on the time horizon. The first instant at 0 hours is equivalent to performing classification with the NN model directly to the state vector obtained from the patient. From there, we perform forecasting up to 40 hours with the DBN model of this state vector and use the results as input for the NN model. We can see that the NN model performs considerably better if we pair it with the DBN to classify the forecasted state of the patients rather than their initial state. As we forecast the state vector of patients further into the future, the NN improves its classification performance monotonically. 

\begin{figure}[!t]\centering
	\includegraphics[width=0.7\columnwidth]{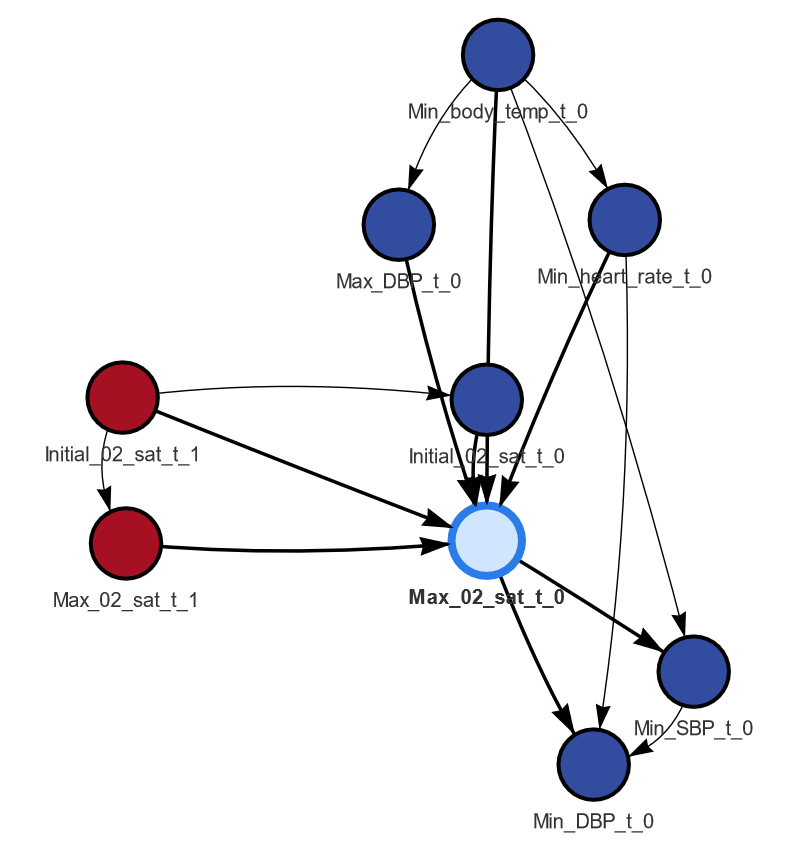}
	\caption{Subset of relevant variables to the forecasting of maximum oxygen saturation (light blue) in the DBN model. The initial and maximum oxygen saturation variables from the last instant (in red) affect the calculation of the next maximum oxygen saturation value. Other variables like body temperature, systolic and diastolic blood pressures and heart rate also influence this value in the forecast.}
	\label{fig:dbn_crop}
\end{figure}

In addition, DBNs perform multivariate inference and are interpretable models. This allows them to offer doctors the forecasted values of any variable in the system as well as the underlying relationships between the rest of variables that led to those results. In the case of relevant values like the oxygen saturation of a patient, which is a good indicator of the state of a patient suffering from respiratory issues, we show an example of the relationships present in the DBN model in Fig. (\ref{fig:dbn_crop}). This subgraph shows the variables directly related with the maximum oxygen saturation registered in a 4 hours interval. We can see the previous maximum value of oxygen saturation from the last instant, which is to be expected due to the autoregressive component of time series. On a similar note, the initial values of oxygen saturation registered serve the model to define the range of the maximum value: lower initial oxygen saturation will likely lead to lower maximums and vice versa. Additionally, we also find in that the body temperature leading to fever, maximum diastolic blood pressure and minimum heart rate are also direct indicators of maximum oxygen saturation and play an important role in its forecasting. Lastly, minimum systolic and diastolic blood pressures are also affected. A lower level of oxygen saturation will cause higher blood pressure, increasing both minimums. This situation is reflected in the fact that this values are child nodes that depend on the current value of oxygen saturation.

\section{Conclusions}\label{sec:concl}

In this work we have presented a hybrid model between DBNs and static classifiers where the state vector recovered from a patient suffering from COVID-19 is used to forecast their future state. This information is then used to assess how severe will their infection be in the future 40 hours based on their current vital signs and blood analysis. This method shows the best performance when combining DBN and NN models. While the NN is capable of discerning whether or not a patient will reach a critical state with better accuracy than the other classifiers, the DBN adds an explainable layer regarding the variables extracted from the patient. This model could help doctors decide whether or not a patient needs further specialized care and allow for a better organization of the resources available in medical centers. Additionally, we offer the code of all our models online for future reference and use.

For future work, this model could be applied in different industrial environments that require forecasting time series and classifying the state of the system. The combination of a generative model that forecasts the state of a system with a classifier model that evaluates this expected future state is a promising framework that could prove useful in applications like remaining useful life estimations. Another possible improvement could be the potential use of the DBN model as a simulator, introducing interventions in the forecasting in order to see the effects that possible actions can have in the expected future. In the medical case, the effects of specific meds or treatments could potentially be reflected in the DBN predictions, and in other industrial cases this could lead to optimizing the expected future based on possible interventions in the initial state.

\section*{Acknowledgements}

This work was partially supported by the Madrid Autonomous Region through the ``MadridDataSpace4Pandemics-CM'' (REACT-EU) project. We are also grateful to the Fundación Jiménez Díaz for providing the data for this work and to the doctors Sara Heili and Lucía Llanos for their valuable insights.

\bibliographystyle{model1-num-names}
\bibliography{preprint_class_DBN_covid}

\begin{thebibliography}{33}
\expandafter\ifx\csname natexlab\endcsname\relax\def\natexlab#1{#1}\fi
\providecommand{\bibinfo}[2]{#2}
\ifx\xfnm\relax \def\xfnm[#1]{\unskip,\space#1}\fi
\bibitem[{Ar{\'e}valo-Lorido et~al.(2022)Ar{\'e}valo-Lorido,
  Carretero-G{\'o}mez, Casas-Rojo, Ant{\'o}n-Santos, Melero-Bermejo,
  L{\'o}pez-Carmona, Palacios, Sanz-C{\'a}novas, Pesqueira-Font{\'a}n, de~la
  Pe{\~n}a-Fern{\'a}ndez et~al.}]{arevalo2022comorb}
\bibinfo{author}{J.~C. Ar{\'e}valo-Lorido},
  \bibinfo{author}{J.~Carretero-G{\'o}mez}, \bibinfo{author}{J.~M. Casas-Rojo},
  \bibinfo{author}{J.~M. Ant{\'o}n-Santos}, \bibinfo{author}{J.~A.
  Melero-Bermejo}, \bibinfo{author}{M.~D. L{\'o}pez-Carmona},
  \bibinfo{author}{L.~C. Palacios}, \bibinfo{author}{J.~Sanz-C{\'a}novas},
  \bibinfo{author}{P.~M. Pesqueira-Font{\'a}n}, \bibinfo{author}{A.~A. de~la
  Pe{\~n}a-Fern{\'a}ndez}, et~al.,
\newblock \bibinfo{title}{The importance of association of comorbidities on
  {COVID-19} outcomes: a machine learning approach},
\newblock \bibinfo{journal}{Current Medical Research and Opinion}
  \bibinfo{volume}{38} (\textbf{\bibinfo{year}{2022}})
  \bibinfo{pages}{501--510}.
\bibitem[{Yu et~al.(2021)Yu, Travaglio, Popovic, Leal, and
  Martins}]{yu2021alzheimer}
\bibinfo{author}{Y.~Yu}, \bibinfo{author}{M.~Travaglio},
  \bibinfo{author}{R.~Popovic}, \bibinfo{author}{N.~S. Leal},
  \bibinfo{author}{L.~M. Martins},
\newblock \bibinfo{title}{{Alzheimer}’s and {Parkinson}’s diseases predict
  different {COVID-19} outcomes: a {UK} {Biobank} study},
\newblock \bibinfo{journal}{Geriatrics} \bibinfo{volume}{6}
  (\textbf{\bibinfo{year}{2021}}) \bibinfo{pages}{10}.
\bibitem[{Ehwerhemuepha et~al.(2021)Ehwerhemuepha, Danioko, Verma, Marano,
  Feaster, Taraman, Moreno, Zheng, Yaghmaei, and
  Chang}]{ehwerhemuepha2021super}
\bibinfo{author}{L.~Ehwerhemuepha}, \bibinfo{author}{S.~Danioko},
  \bibinfo{author}{S.~Verma}, \bibinfo{author}{R.~Marano},
  \bibinfo{author}{W.~Feaster}, \bibinfo{author}{S.~Taraman},
  \bibinfo{author}{T.~Moreno}, \bibinfo{author}{J.~Zheng},
  \bibinfo{author}{E.~Yaghmaei}, \bibinfo{author}{A.~Chang},
\newblock \bibinfo{title}{A super learner ensemble of 14 statistical learning
  models for predicting {COVID-19} severity among patients with cardiovascular
  conditions},
\newblock \bibinfo{journal}{Intelligence-Based Medicine} \bibinfo{volume}{5}
  (\textbf{\bibinfo{year}{2021}}) \bibinfo{pages}{100030}.
\bibitem[{Momeni-Boroujeni et~al.(2021)Momeni-Boroujeni, Mendoza, Stopard,
  Lambert, and Zuretti}]{momeni2021dynamic}
\bibinfo{author}{A.~Momeni-Boroujeni}, \bibinfo{author}{R.~Mendoza},
  \bibinfo{author}{I.~J. Stopard}, \bibinfo{author}{B.~Lambert},
  \bibinfo{author}{A.~Zuretti},
\newblock \bibinfo{title}{A dynamic {Bayesian} model for identifying
  high-mortality risk in hospitalized {COVID-19} patients},
\newblock \bibinfo{journal}{Infectious Disease Reports} \bibinfo{volume}{13}
  (\textbf{\bibinfo{year}{2021}}) \bibinfo{pages}{239--250}.
\bibitem[{Wu et~al.(2020)Wu, Yang, Xie, Woodruff, Rao, Guiot, Frix, Louis,
  Moutschen, Li et~al.}]{wu2020development}
\bibinfo{author}{G.~Wu}, \bibinfo{author}{P.~Yang}, \bibinfo{author}{Y.~Xie},
  \bibinfo{author}{H.~C. Woodruff}, \bibinfo{author}{X.~Rao},
  \bibinfo{author}{J.~Guiot}, \bibinfo{author}{A.-N. Frix},
  \bibinfo{author}{R.~Louis}, \bibinfo{author}{M.~Moutschen},
  \bibinfo{author}{J.~Li}, et~al.,
\newblock \bibinfo{title}{Development of a clinical decision support system for
  severity risk prediction and triage of {COVID-19} patients at hospital
  admission: an international multicentre study},
\newblock \bibinfo{journal}{European Respiratory Journal} \bibinfo{volume}{56}
  (\textbf{\bibinfo{year}{2020}}).
\bibitem[{Xiong et~al.(2020)Xiong, Zhang, Watson, Sundin, Bufford, Zoller,
  Shamshoian, Suchard, and Ramirez}]{xiong2020pseudo}
\bibinfo{author}{D.~Xiong}, \bibinfo{author}{L.~Zhang}, \bibinfo{author}{G.~L.
  Watson}, \bibinfo{author}{P.~Sundin}, \bibinfo{author}{T.~Bufford},
  \bibinfo{author}{J.~A. Zoller}, \bibinfo{author}{J.~Shamshoian},
  \bibinfo{author}{M.~A. Suchard}, \bibinfo{author}{C.~M. Ramirez},
\newblock \bibinfo{title}{Pseudo-likelihood based logistic regression for
  estimating {COVID-19} infection and case fatality rates by gender, race, and
  age in {California}},
\newblock \bibinfo{journal}{Epidemics} \bibinfo{volume}{33}
  (\textbf{\bibinfo{year}{2020}}) \bibinfo{pages}{100418}.
\bibitem[{Berenguer et~al.(2021)Berenguer, Borobia, Ryan,
  Rodr{\'\i}guez-Ba{\~n}o, Bell{\'o}n, Jarr{\'\i}n, Carratal{\`a}, Pach{\'o}n,
  Carcas, Yllescas, and Arribas}]{berenguer920logistic}
\bibinfo{author}{J.~Berenguer}, \bibinfo{author}{A.~M. Borobia},
  \bibinfo{author}{P.~Ryan}, \bibinfo{author}{J.~Rodr{\'\i}guez-Ba{\~n}o},
  \bibinfo{author}{J.~M. Bell{\'o}n}, \bibinfo{author}{I.~Jarr{\'\i}n},
  \bibinfo{author}{J.~Carratal{\`a}}, \bibinfo{author}{J.~Pach{\'o}n},
  \bibinfo{author}{A.~J. Carcas}, \bibinfo{author}{M.~Yllescas},
  \bibinfo{author}{J.~R. Arribas},
\newblock \bibinfo{title}{Development and validation of a prediction model for
  30-day mortality in hospitalised patients with {COVID-19}: the {COVID-19}
  {SEIMC} score},
\newblock \bibinfo{journal}{Thorax} \bibinfo{volume}{76}
  (\textbf{\bibinfo{year}{2021}}) \bibinfo{pages}{920--929}.
\bibitem[{Pinter et~al.(2020)Pinter, Felde, Mosavi, Ghamisi, and
  Gloaguen}]{pinter2020covid}
\bibinfo{author}{G.~Pinter}, \bibinfo{author}{I.~Felde},
  \bibinfo{author}{A.~Mosavi}, \bibinfo{author}{P.~Ghamisi},
  \bibinfo{author}{R.~Gloaguen},
\newblock \bibinfo{title}{{COVID-19} pandemic prediction for {Hungary}; a
  hybrid machine learning approach},
\newblock \bibinfo{journal}{Mathematics} \bibinfo{volume}{8}
  (\textbf{\bibinfo{year}{2020}}) \bibinfo{pages}{890}.
\bibitem[{Dhamodharavadhani et~al.(2020)Dhamodharavadhani, Rathipriya, and
  Chatterjee}]{dhamodharavadhani2020covid}
\bibinfo{author}{S.~Dhamodharavadhani}, \bibinfo{author}{R.~Rathipriya},
  \bibinfo{author}{J.~M. Chatterjee},
\newblock \bibinfo{title}{{COVID-19} mortality rate prediction for {India}
  using statistical neural network models},
\newblock \bibinfo{journal}{Frontiers in Public Health} \bibinfo{volume}{8}
  (\textbf{\bibinfo{year}{2020}}) \bibinfo{pages}{441}.
\bibitem[{Aznar-Gimeno et~al.(2021)Aznar-Gimeno, Esteban, Labata-Lezaun, del
  Hoyo-Alonso, Abadia-Gallego, Pa{\~n}o-Pardo, Esquillor-Rodrigo, Lanas, and
  Serrano}]{aznar2021clinical}
\bibinfo{author}{R.~Aznar-Gimeno}, \bibinfo{author}{L.~M. Esteban},
  \bibinfo{author}{G.~Labata-Lezaun}, \bibinfo{author}{R.~del Hoyo-Alonso},
  \bibinfo{author}{D.~Abadia-Gallego}, \bibinfo{author}{J.~R. Pa{\~n}o-Pardo},
  \bibinfo{author}{M.~J. Esquillor-Rodrigo}, \bibinfo{author}{{\'A}.~Lanas},
  \bibinfo{author}{M.~T. Serrano},
\newblock \bibinfo{title}{A clinical decision web to predict {ICU} admission or
  death for patients hospitalised with {COVID-19} using machine learning
  algorithms},
\newblock \bibinfo{journal}{International Journal of Environmental Research and
  Public Health} \bibinfo{volume}{18} (\textbf{\bibinfo{year}{2021}})
  \bibinfo{pages}{8677}.
\bibitem[{Kianfar et~al.(2022)Kianfar, Mesgari, Mollalo, and
  Kaveh}]{kianfar2022spatio}
\bibinfo{author}{N.~Kianfar}, \bibinfo{author}{M.~S. Mesgari},
  \bibinfo{author}{A.~Mollalo}, \bibinfo{author}{M.~Kaveh},
\newblock \bibinfo{title}{Spatio-temporal modeling of {COVID-19} prevalence and
  mortality using artificial neural network algorithms},
\newblock \bibinfo{journal}{Spatial and Spatio-temporal Epidemiology}
  \bibinfo{volume}{40} (\textbf{\bibinfo{year}{2022}}) \bibinfo{pages}{100471}.
\bibitem[{Cornelius et~al.(2021)Cornelius, Akman, and
  Hrozencik}]{cornelius2021rf}
\bibinfo{author}{E.~Cornelius}, \bibinfo{author}{O.~Akman},
  \bibinfo{author}{D.~Hrozencik},
\newblock \bibinfo{title}{{COVID-19} mortality prediction using machine
  learning-integrated {Random} {Forest} algorithm under varying patient
  frailty},
\newblock \bibinfo{journal}{Mathematics} \bibinfo{volume}{9}
  (\textbf{\bibinfo{year}{2021}}) \bibinfo{pages}{2043}.
\bibitem[{Pourhomayoun and Shakibi(2021)}]{pourhomayoun2021rf}
\bibinfo{author}{M.~Pourhomayoun}, \bibinfo{author}{M.~Shakibi},
\newblock \bibinfo{title}{Predicting mortality risk in patients with {COVID-19}
  using machine learning to help medical decision-making},
\newblock \bibinfo{journal}{Smart Health} \bibinfo{volume}{20}
  (\textbf{\bibinfo{year}{2021}}) \bibinfo{pages}{100178}.
\bibitem[{Tezza et~al.(2021)Tezza, Lorenzoni, Azzolina, Barbar, Leone, and
  Gregori}]{tezza2021rf}
\bibinfo{author}{F.~Tezza}, \bibinfo{author}{G.~Lorenzoni},
  \bibinfo{author}{D.~Azzolina}, \bibinfo{author}{S.~Barbar},
  \bibinfo{author}{L.~A.~C. Leone}, \bibinfo{author}{D.~Gregori},
\newblock \bibinfo{title}{Predicting in-hospital mortality of patients with
  {COVID-19} using machine learning techniques},
\newblock \bibinfo{journal}{Journal of Personalized Medicine}
  \bibinfo{volume}{11} (\textbf{\bibinfo{year}{2021}}) \bibinfo{pages}{343}.
\bibitem[{Bertsimas et~al.(2020)Bertsimas, Lukin, Mingardi, Nohadani,
  Orfanoudaki, Stellato, Wiberg, Gonzalez-Garcia, Parra-Calder{\'o}n, Robinson
  et~al.}]{bertsimas2020xgb}
\bibinfo{author}{D.~Bertsimas}, \bibinfo{author}{G.~Lukin},
  \bibinfo{author}{L.~Mingardi}, \bibinfo{author}{O.~Nohadani},
  \bibinfo{author}{A.~Orfanoudaki}, \bibinfo{author}{B.~Stellato},
  \bibinfo{author}{H.~Wiberg}, \bibinfo{author}{S.~Gonzalez-Garcia},
  \bibinfo{author}{C.~L. Parra-Calder{\'o}n}, \bibinfo{author}{K.~Robinson},
  et~al.,
\newblock \bibinfo{title}{{COVID-19} mortality risk assessment: an
  international multi-center study},
\newblock \bibinfo{journal}{PLOS ONE} \bibinfo{volume}{15}
  (\textbf{\bibinfo{year}{2020}}) \bibinfo{pages}{e0243262}.
\bibitem[{Vaid et~al.(2020)Vaid, Somani, Russak, De~Freitas, Chaudhry,
  Paranjpe, Johnson, Lee, Miotto, Richter et~al.}]{vaid2020xgb}
\bibinfo{author}{A.~Vaid}, \bibinfo{author}{S.~Somani},
  \bibinfo{author}{A.~Russak}, \bibinfo{author}{J.~De~Freitas},
  \bibinfo{author}{F.~Chaudhry}, \bibinfo{author}{I.~Paranjpe},
  \bibinfo{author}{K.~Johnson}, \bibinfo{author}{S.~Lee},
  \bibinfo{author}{R.~Miotto}, \bibinfo{author}{F.~Richter}, et~al.,
\newblock \bibinfo{title}{Machine learning to predict mortality and critical
  events in {COVID-19} positive {New} {York} {City} patients: a cohort study.},
\newblock \bibinfo{journal}{Journal of Medical Internet Research}
  (\textbf{\bibinfo{year}{2020}}).
\bibitem[{Yadaw et~al.(2020)Yadaw, Li, Bose, Iyengar, Bunyavanich, and
  Pandey}]{yadaw2020xgb}
\bibinfo{author}{A.~S. Yadaw}, \bibinfo{author}{Y.-c. Li},
  \bibinfo{author}{S.~Bose}, \bibinfo{author}{R.~Iyengar},
  \bibinfo{author}{S.~Bunyavanich}, \bibinfo{author}{G.~Pandey},
\newblock \bibinfo{title}{Clinical predictors of {COVID-19} mortality},
\newblock \bibinfo{journal}{MedRxiv}  (\textbf{\bibinfo{year}{2020}}).
\bibitem[{Fenton et~al.(2021)Fenton, McLachlan, Lucas, Dube, Hitman, Osman,
  Kyrimi, and Neil}]{fenton2021bayesian}
\bibinfo{author}{N.~E. Fenton}, \bibinfo{author}{S.~McLachlan},
  \bibinfo{author}{P.~Lucas}, \bibinfo{author}{K.~Dube}, \bibinfo{author}{G.~A.
  Hitman}, \bibinfo{author}{M.~Osman}, \bibinfo{author}{E.~Kyrimi},
  \bibinfo{author}{M.~Neil},
\newblock \bibinfo{title}{A {Bayesian} network model for personalised
  {COVID-19} risk assessment and contact tracing},
\newblock \bibinfo{journal}{MedRxiv}  (\textbf{\bibinfo{year}{2021}})
  \bibinfo{pages}{2020--07}.
\bibitem[{Vepa et~al.(2021)Vepa, Saleem, Rakhshan, Daneshkhah, Sedighi,
  Shohaimi, Omar, Salari, Chatrabgoun, Dharmaraj et~al.}]{vepa2021bn}
\bibinfo{author}{A.~Vepa}, \bibinfo{author}{A.~Saleem},
  \bibinfo{author}{K.~Rakhshan}, \bibinfo{author}{A.~Daneshkhah},
  \bibinfo{author}{T.~Sedighi}, \bibinfo{author}{S.~Shohaimi},
  \bibinfo{author}{A.~Omar}, \bibinfo{author}{N.~Salari},
  \bibinfo{author}{O.~Chatrabgoun}, \bibinfo{author}{D.~Dharmaraj}, et~al.,
\newblock \bibinfo{title}{Using machine learning algorithms to develop a
  clinical decision-making tool for {COVID-19} inpatients},
\newblock \bibinfo{journal}{International Journal of Environmental Research and
  Public Health} \bibinfo{volume}{18} (\textbf{\bibinfo{year}{2021}})
  \bibinfo{pages}{6228}.
\bibitem[{Pezoulas et~al.(2021)Pezoulas, Kourou, Papaloukas, Triantafyllia,
  Lampropoulou, Siouti, Papadaki, Salagianni, Koukaki, Rovina
  et~al.}]{pezoulas2021ts}
\bibinfo{author}{V.~C. Pezoulas}, \bibinfo{author}{K.~D. Kourou},
  \bibinfo{author}{C.~Papaloukas}, \bibinfo{author}{V.~Triantafyllia},
  \bibinfo{author}{V.~Lampropoulou}, \bibinfo{author}{E.~Siouti},
  \bibinfo{author}{M.~Papadaki}, \bibinfo{author}{M.~Salagianni},
  \bibinfo{author}{E.~Koukaki}, \bibinfo{author}{N.~Rovina}, et~al.,
\newblock \bibinfo{title}{A multimodal approach for the risk prediction of
  intensive care and mortality in patients with {COVID-19}},
\newblock \bibinfo{journal}{Diagnostics} \bibinfo{volume}{12}
  (\textbf{\bibinfo{year}{2021}}) \bibinfo{pages}{56}.
\bibitem[{Villegas et~al.(2021)Villegas, Gonzalez-Agirre,
  Guti{\'e}rrez-Fandi{\~n}o, Armengol-Estap{\'e}, Carrino, Fern{\'a}ndez,
  Soares, Serrano, Pedrera, Garc{\'\i}a et~al.}]{villegas2021ts}
\bibinfo{author}{M.~Villegas}, \bibinfo{author}{A.~Gonzalez-Agirre},
  \bibinfo{author}{A.~Guti{\'e}rrez-Fandi{\~n}o},
  \bibinfo{author}{J.~Armengol-Estap{\'e}}, \bibinfo{author}{C.~P. Carrino},
  \bibinfo{author}{D.~P. Fern{\'a}ndez}, \bibinfo{author}{F.~Soares},
  \bibinfo{author}{P.~Serrano}, \bibinfo{author}{M.~Pedrera},
  \bibinfo{author}{N.~Garc{\'\i}a}, et~al.,
\newblock \bibinfo{title}{Predicting the evolution of {COVID-19} mortality
  risk: a recurrent neural network approach},
\newblock \bibinfo{journal}{MedRxiv}  (\textbf{\bibinfo{year}{2021}})
  \bibinfo{pages}{2020--12}.
\bibitem[{Momeni-Boroujeni et~al.(2021)Momeni-Boroujeni, Mendoza, Stopard,
  Lambert, and Zuretti}]{momeni2021ts}
\bibinfo{author}{A.~Momeni-Boroujeni}, \bibinfo{author}{R.~Mendoza},
  \bibinfo{author}{I.~J. Stopard}, \bibinfo{author}{B.~Lambert},
  \bibinfo{author}{A.~Zuretti},
\newblock \bibinfo{title}{A dynamic {Bayesian} model for identifying
  high-mortality risk in hospitalized {COVID-19} patients},
\newblock \bibinfo{journal}{Infectious Disease Reports} \bibinfo{volume}{13}
  (\textbf{\bibinfo{year}{2021}}) \bibinfo{pages}{239--250}.
\bibitem[{Koller and Friedman(2009)}]{koller2009}
\bibinfo{author}{D.~Koller}, \bibinfo{author}{N.~Friedman},
  \bibinfo{title}{Probabilistic Graphical Models: Principles and Techniques},
  \bibinfo{publisher}{The MIT Press}, \textbf{\bibinfo{year}{2009}}.
\bibitem[{Fan et~al.(2020)Fan, Tu, Zhou, Liu, Wang, Song, Gu, Wang, Wei, Li
  et~al.}]{fan2020score}
\bibinfo{author}{G.~Fan}, \bibinfo{author}{C.~Tu}, \bibinfo{author}{F.~Zhou},
  \bibinfo{author}{Z.~Liu}, \bibinfo{author}{Y.~Wang},
  \bibinfo{author}{B.~Song}, \bibinfo{author}{X.~Gu},
  \bibinfo{author}{Y.~Wang}, \bibinfo{author}{Y.~Wei}, \bibinfo{author}{H.~Li},
  et~al.,
\newblock \bibinfo{title}{Comparison of severity scores for {COVID-19} patients
  with pneumonia: a retrospective study},
\newblock \bibinfo{journal}{European Respiratory Journal} \bibinfo{volume}{56}
  (\textbf{\bibinfo{year}{2020}}).
\bibitem[{Keogh and Pazzani(2002)}]{keogh2002learning}
\bibinfo{author}{E.~J. Keogh}, \bibinfo{author}{M.~J. Pazzani},
\newblock \bibinfo{title}{Learning the structure of augmented {Bayesian}
  classifiers},
\newblock \bibinfo{journal}{International Journal on Artificial Intelligence
  Tools} \bibinfo{volume}{11} (\textbf{\bibinfo{year}{2002}})
  \bibinfo{pages}{587--601}.
\bibitem[{Chen et~al.(2022)Chen, He, Benesty, Khotilovich, Tang, Cho, Chen,
  Mitchell, Cano, Zhou, Li, Xie, Lin, Geng, Li, and Yuan}]{chen2022xgboost}
\bibinfo{author}{T.~Chen}, \bibinfo{author}{T.~He},
  \bibinfo{author}{M.~Benesty}, \bibinfo{author}{V.~Khotilovich},
  \bibinfo{author}{Y.~Tang}, \bibinfo{author}{H.~Cho},
  \bibinfo{author}{K.~Chen}, \bibinfo{author}{R.~Mitchell},
  \bibinfo{author}{I.~Cano}, \bibinfo{author}{T.~Zhou},
  \bibinfo{author}{M.~Li}, \bibinfo{author}{J.~Xie}, \bibinfo{author}{M.~Lin},
  \bibinfo{author}{Y.~Geng}, \bibinfo{author}{Y.~Li},
  \bibinfo{author}{J.~Yuan}, \bibinfo{title}{{xgboost}: {Extreme} gradient
  boosting}, \textbf{\bibinfo{year}{2022}}. \bibinfo{note}{R package version
  1.6.0.1}.
\bibitem[{Meyer et~al.(2022)Meyer, Dimitriadou, Hornik, Weingessel, and
  Leisch}]{meyer2022e1071}
\bibinfo{author}{D.~Meyer}, \bibinfo{author}{E.~Dimitriadou},
  \bibinfo{author}{K.~Hornik}, \bibinfo{author}{A.~Weingessel},
  \bibinfo{author}{F.~Leisch}, \bibinfo{title}{{e1071}: Misc functions of the
  {Department} of {Statistics}, {Probability} {Theory} {Group} {(Formerly:
  E1071), TU Wien}}, \textbf{\bibinfo{year}{2022}}. \bibinfo{note}{R package
  version 1.7-12}.
\bibitem[{Allaire and Chollet(2022)}]{allaire2022keras}
\bibinfo{author}{J.~Allaire}, \bibinfo{author}{F.~Chollet},
  \bibinfo{title}{{keras}: R Interface to {`Keras'}},
  \textbf{\bibinfo{year}{2022}}. \bibinfo{note}{R package version 2.9.0}.
\bibitem[{{Mihaljevi'{c}} et~al.(2018){Mihaljevi'{c}}, Bielza, and
  {Larra\~naga}}]{mihaljevic2018bnclassify}
\bibinfo{author}{B.~{Mihaljevi'{c}}}, \bibinfo{author}{C.~Bielza},
  \bibinfo{author}{P.~{Larra\~naga}},
\newblock \bibinfo{title}{{bnclassify}: {Learning} {Bayesian} network
  classifiers},
\newblock \bibinfo{journal}{The R Journal} \bibinfo{volume}{10}
  (\textbf{\bibinfo{year}{2018}}) \bibinfo{pages}{455--468}.
\bibitem[{Mullen et~al.(2011)Mullen, Ardia, Gil, Windover, and
  Cline}]{mullen2011deoptim}
\bibinfo{author}{K.~Mullen}, \bibinfo{author}{D.~Ardia},
  \bibinfo{author}{D.~Gil}, \bibinfo{author}{D.~Windover},
  \bibinfo{author}{J.~Cline},
\newblock \bibinfo{title}{{DEoptim}: An {R} package for global optimization by
  {D}ifferential {E}volution},
\newblock \bibinfo{journal}{Journal of Statistical Software}
  \bibinfo{volume}{40} (\textbf{\bibinfo{year}{2011}}) \bibinfo{pages}{1--26}.
\bibitem[{Tharwat(2021)}]{tharwat2021classification}
\bibinfo{author}{A.~Tharwat},
\newblock \bibinfo{title}{Classification assessment methods},
\newblock \bibinfo{journal}{Applied computing and informatics}
  \bibinfo{volume}{17} (\textbf{\bibinfo{year}{2021}})
  \bibinfo{pages}{168--192}.
\bibitem[{Chawla et~al.(2002)Chawla, Bowyer, Hall, and
  Kegelmeyer}]{chawla2002smote}
\bibinfo{author}{N.~V. Chawla}, \bibinfo{author}{K.~W. Bowyer},
  \bibinfo{author}{L.~O. Hall}, \bibinfo{author}{W.~P. Kegelmeyer},
\newblock \bibinfo{title}{{SMOTE}: synthetic minority over-sampling technique},
\newblock \bibinfo{journal}{Journal of artificial intelligence research}
  \bibinfo{volume}{16} (\textbf{\bibinfo{year}{2002}})
  \bibinfo{pages}{321--357}.
\bibitem[{Torgo(2010)}]{torgo2010dmwr}
\bibinfo{author}{L.~Torgo}, \bibinfo{title}{Data Mining with R, learning with
  case studies}, \bibinfo{publisher}{Chapman and Hall/CRC},
  \textbf{\bibinfo{year}{2010}}.

\end{thebibliography}

\end{document}